\documentclass{article}

\usepackage{microtype}
\usepackage{graphicx}
\usepackage{subcaption}
\usepackage{booktabs} 
\usepackage{hyperref}

\usepackage[accepted]{icml2026}
\usepackage{amsmath}
\usepackage{amssymb}
\usepackage{mathtools}
\usepackage{amsthm}
\usepackage[capitalize,noabbrev]{cleveref}

\newcommand{\method}{SLIP-RS}

\theoremstyle{plain}

\theoremstyle{definition}

\theoremstyle{remark}

\usepackage[textsize=tiny]{todonotes}
\usepackage{enumitem}
\usepackage{booktabs}
\usepackage{multirow}
\usepackage{makecell}

\icmltitlerunning{\method{}: Structured-Attribute Language-Image Pre-Training for Remote Sensing Object Detection}

\begin{document}

\twocolumn[
  \icmltitle{\method{}: Structured-Attribute Language-Image Pre-Training for \\
  Remote Sensing Object Detection}

  \icmlsetsymbol{equal}{*}

  \begin{icmlauthorlist}
    \icmlauthor{Chenxu Wang}{1}
    \icmlauthor{Yuxuan Li}{1}
    \icmlauthor{Yunheng Li}{1}
    \icmlauthor{Xiang Li}{1,2}
    \icmlauthor{Jingyuan Xia}{3}
    \icmlauthor{Qibin Hou}{1,2}
  \end{icmlauthorlist}

  \icmlaffiliation{1}{VCIP, CS, Nankai University}
  \icmlaffiliation{2}{NKIARI, Shenzhen Futian}
  \icmlaffiliation{3}{National University of Defense Technology}

  \icmlcorrespondingauthor{Qibin Hou}{houqb@nankai.edu.cn}

  \icmlkeywords{Machine Learning, ICML}

  \vskip 0.3in
]

\printAffiliationsAndNotice{}

\begin{abstract}
Existing language-image pre-training for remote sensing object detection is constrained by Monolithic Label Learning, which relies on exhaustively enumerating open-set categories via black-box data to acquire fine-grained representations, creating a dependency incompatible with the domain's inherent data scarcity. 
To transcend this bottleneck, we propose \method{}, establishing a Structured-Attribute Decoupling Paradigm that maps the open-ended category space into a finite, physically meaningful attribute space, unlocking fine-grained discriminability via explicit structural logic. 
This paradigm is realized via two technical pillars: 
(1) Structured-Attribute Contrastive Learning, which enforces the learning of decoupled intrinsic visual logic via combinatorial attribute augmentation; 
and (2) Conformal Attribute Reliability Engine, which leverages conformal prediction theory to rigorously distill high-fidelity supervision from noisy sources, yielding RS-Attribute-15M, the largest dataset with over 15 million attribute annotations. 
Extensive experiments demonstrate that \method{} establishes unprecedented performance in fine-grained detection and cross-domain generalization. 
Code: \url{https://github.com/facias914/SLIP-RS}. 

\end{abstract}

\section{Introduction}

\begin{figure}[t]
  \centering
  \setlength{\abovecaptionskip}{2pt}
  \includegraphics[width=\linewidth]{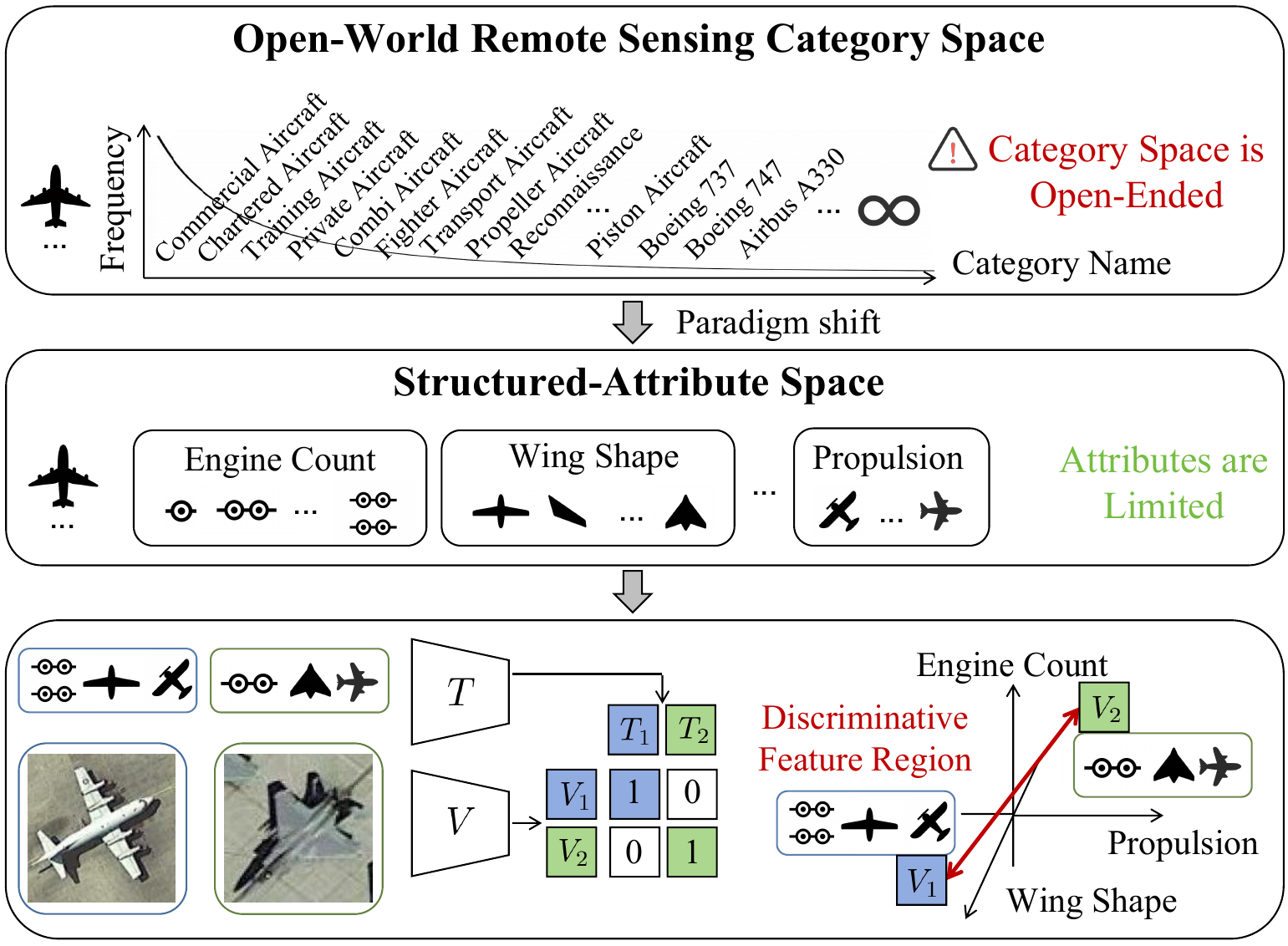}
  \caption{Structured-Attribute Decoupling Paradigm that decomposes objects into finite structural attributes, enabling scalable and discriminative fine-grained representation learning. }
  \label{figure1}
\end{figure}

Large-scale pre-training for remote sensing object detection has emerged as a pivotal research direction, aiming to propel the field from coarse-grained localization to fine-grained semantic understanding. 
This paradigm is essential to transcend the dual bottlenecks of scarce annotated data and limited category coverage inherent in closed-set detectors~\cite{yang2021r3det,li2024lsknet,cai2024poly}.
Ideally, such a pre-training framework should endow the model with robust and transferable feature representations. These representations enable the model to not only encompass a broad spectrum of fine-grained categories with superior domain generalization during pre-training but also serve as a solid foundation that flexibly adapts to diverse novel categories and scenarios via downstream fine-tuning. 
However, realizing this vision faces a fundamental challenge: How can we effectively expand the perception boundaries of fine-grained categories from an open-ended category space where detailed supervision is notoriously scarce? 

To bridge the category expansion bottleneck in remote sensing, mainstream research~\cite{pan2025locate,huang2025openrsd} predominantly adopts the Open-Vocabulary Learning paradigm from the general vision domain~\cite{li2022grounded,liu2024grounding}. 
Inspired by CLIP~\cite{radford2021learning}, this paradigm leverages region-level image–language contrastive learning to map visual features into the linguistic space, aiming to extending closed-set detectors~\cite{Dai_2021_CVPR,zhang2022dino} to open-set scenarios through the semantically rich language. 
However, we contend that this approach fundamentally remains trapped in the dogma of Monolithic Label Learning. 
It merely shifts from the mechanical memorization of discrete category IDs to the holistic matching of continuous category name embeddings, without truly decoupling the semantic structure. 

The above paradigm encounters two insurmountable barriers in the context of remote sensing fine-grained perception. 
First is the data-driven scalability bottleneck. 
This paradigm relies on \textit{black-box} image-text pairs to exhaustively enumerate the open-ended category space, thereby implicitly learning fine-grained representations. 
However, the remote sensing domain naturally lacks comparable public data sources with high semantic density, depriving the model of sufficient supervisory signals for learning fine-grained representations. 
Second is the loss of discriminability caused by Semantic Binding. 
The model continues to treat each category (e.g., \textit{Boeing 747}) as an inextricable semantic black dot, rather than a set of independent and recombinable visual attributes (e.g., \textit{four engines + swept wings}). 
This neglect of object structure leads to a recognition blind spot, where the model struggles to capture key discriminative features when distinguishing targets with subtle attribute differences. 

To address these dilemmas, we argue that the essence of object semantics should not be confined to a single label but be reconceptualized as a structured combination of attributes (as shown in Figure~\ref{figure1}).
Inspired by this, we propose \method{} (\textbf{S}tructured-Attribute \textbf{L}anguage-\textbf{I}mage \textbf{P}re-training for \textbf{R}emote \textbf{S}ensing), whose core contribution lies in establishing an Structured-Attribute Decoupling Paradigm. 
Instead of mapping objects to static, indivisible monolithic labels, we decompose them into a set of orthogonal and mutually exclusive semantic primitives (e.g., \textit{engine count} and \textit{wing shape}). 
This paradigm shift fundamentally reshapes the model's representation space. 
On one hand, it endows the model with extreme fine-grained discriminability, enabling it to precisely distinguish visually similar targets through explicit attribute differences (e.g., differentiating engine counts). 
On the other hand, by aligning with a physically meaningful attribute space, the model acquires robust representations rich in structural priors. 
These attribute-level priors transcend simple data memorization, constituting a feature foundation with immense transfer potential. 
This allows the model to leverage its acute perception of geometric structures and fine-grained features to demonstrate superior performance upper bounds during downstream fine-tuning. 

To substantiate this paradigm, we design two core technical pillars. 
The first is the Structured-Attribute Contrastive Learning (SACL) mechanism. 
Through structured-attribute dictionary construction and randomized combinatorial augmentation, SACL enforces the learning of independent, decoupled attribute representations.
This mechanism explicitly prevents shortcut learning based on linguistic templates, ensuring that the model comprehends intrinsic visual logic rather than mere rote memorization. 
The second is the Conformal Attribute Reliability Engine (CARE). 
To address the scarcity of grounding-level attribute supervision, CARE operates via a progressive Tune-Calibrate-Scale pipeline. 
It first fine-tunes a specialized fine-grained teacher model to generate candidate attribute annotations, then critically integrates Conformal Prediction theory~\cite{shafer2008tutorial} during calibration. 
By replacing heuristic thresholding with rigorous statistical boundaries that strictly control the False Discovery Rate, CARE effectively filters noise labels with theoretical guarantees. 
This engine enables the curation of RS-Attribute-15M (comprising over 15 million instance-level attribute annotations), serving as the robust data fuel for learning fine-grained attribute representations. 

Extensive experiments validate \method{}'s superiority over other methods across fine-grained category and attribute detection, cross-domain generalization, and downstream fine-tuning tasks. 
These results confirm that Structured-Attribute Language-Image Pre-Training is a vital pathway for building scalable remote sensing foundation models. 
In summary, our main contributions are: 
\begin{itemize} 
    \item We propose \method{}, establishing a novel Structured-Attribute Decoupling Paradigm that shifts from monolithic labeling to compositional semantic primitives, fundamentally enhancing fine-grained discriminability. 
    \item We design SACL, a Structured-Attribute Contrastive Learning strategy that enforces intrinsic visual logic learning via combinatorial attribute augmentation. 
    \item We introduce the CARE engine to curate RS-Attribute-15M, the first detection dataset with attribute annotations serving as a robust foundation for learning fine-grained attribute representations. 
\end{itemize}

\section{Related Work}

\subsection{Remote Sensing Object Detection}
Benefiting from the flourishing development of deep learning in the general domain~\cite{ren2016faster,redmon2016you,lin2017focal,carion2020end} and the establishment of remote sensing object detection benchmarks~\cite{xia2018dota,sun2022fair1m}, generic detection frameworks have been extensively adapted to the remote sensing domain. 
To address intrinsic challenges such as extreme variations in scale and aspect ratio, arbitrary object orientation, dense object clustering, and high annotation costs, existing research has established a comprehensive array of solutions. 
Specifically, researchers have significantly enhanced performance by innovating in network architecture~\cite{han2021redet,sm3det}, designing advanced label assignment strategies~\cite{xu2022rfla,hou2022shape,xu2023dynamic}, optimizing regression loss functions~\cite{yang2021rethinking,yang2021learning}, and employing label-efficient learning strategies~\cite{hua2023sood,wang2025multi} that leverage unlabeled data for semi-supervised training or learn rotated box regression from low-cost horizontal box~\cite{yang2023h2rbox,yu2023h2rboxv2} and point annotations~\cite{yu2024point2rbox,yu2025point2rbox}. 
Nevertheless, beneath these advancements, most methods treats recognition as discrete ID classification within a static taxonomy, they collapse diverse instances into semantic black dots, inducing a semantic binding that prevents learning discriminative fine-grained representations under data scarcity. 

\subsection{Contrastive Language-Image Pre-training}
The field of computer vision has witnessed a paradigm shift from closed-set label supervision to open-ended language supervision. 
Pioneering works such as CLIP~\cite{radford2021learning}, ALIGN~\cite{jia2021scaling}, and SigLIP~\cite{zhai2023sigmoid} revolutionized visual recognition by introducing contrastive learning on image-text pairs. 
By leveraging the rich semantic information inherent in language modality, these models established a robust foundation for fine-grained discriminative representations learning. 
Inspired by this paradigm, subsequent research~\cite{li2022grounded,liu2024grounding,yao2022detclip,cheng2024yolo} has extended this capability to dense prediction tasks, giving rise to Open-Vocabulary Object Detection. 
Instead of relying on a fixed classifier, these methods align region-level visual features with text embeddings derived from language encoders (e.g., BERT~\cite{devlin2019bert} or CLIP text tower) to leverage the boundless semantics of language. 
Propelled by data engines operating on black-box scale image-text corpora, these frameworks derive robust open-vocabulary capabilities and the emergent power to detect novel categories. 

This paradigm has been extensively adapted to remote sensing, yet confronts unique domain-specific hurdles. 
Efforts such as GRAFT~\cite{mall2023remote}, SkyScript~\cite{wang2024skyscript}, and RemoteCLIP~\cite{liu2024remoteclip} have attempted to replicate this paradigm by constructing domain-specific image-text pairs via ground-satellite alignment or geo-spatial tagging, and ScoreRS~\cite{muhtar2025quality} further introduces preference-based scoring to filter these data for higher quality. 
However, unlike the web-scale corpora in general vision that offer exhaustive fine-grained coverage, these remote sensing datasets are predominantly anchored in coarse-grained alignment. 
The available textual descriptions (e.g., \textit{an aerial view of an aircraft}) inherently fail to capture the object attributes (e.g., \textit{wing shape}) required for fine-grained recognition. 
Meanwhile, remote sensing open-vocabulary object detection confronts an identical bottleneck. 
Although recent approaches~\cite{li2024toward,pan2025locate,huang2025openrsd} attempt to augment semantics via Vision-Language Models~\cite{chen2024internvl} or visual prompts, they remain confined to coarse-grained category expansion. 
Deprived of the massive attribute-rich data needed to implicitly learn fine-grained details~\cite{li2025advancing,li2025anchoropt}, they struggle to penetrate the semantic binding of objects, compromising their utility in learning fine-grained representations.

\begin{figure*}[t] %
    \footnotesize
    \centering
    \setlength{\abovecaptionskip}{2pt}
    \includegraphics[width=\linewidth]{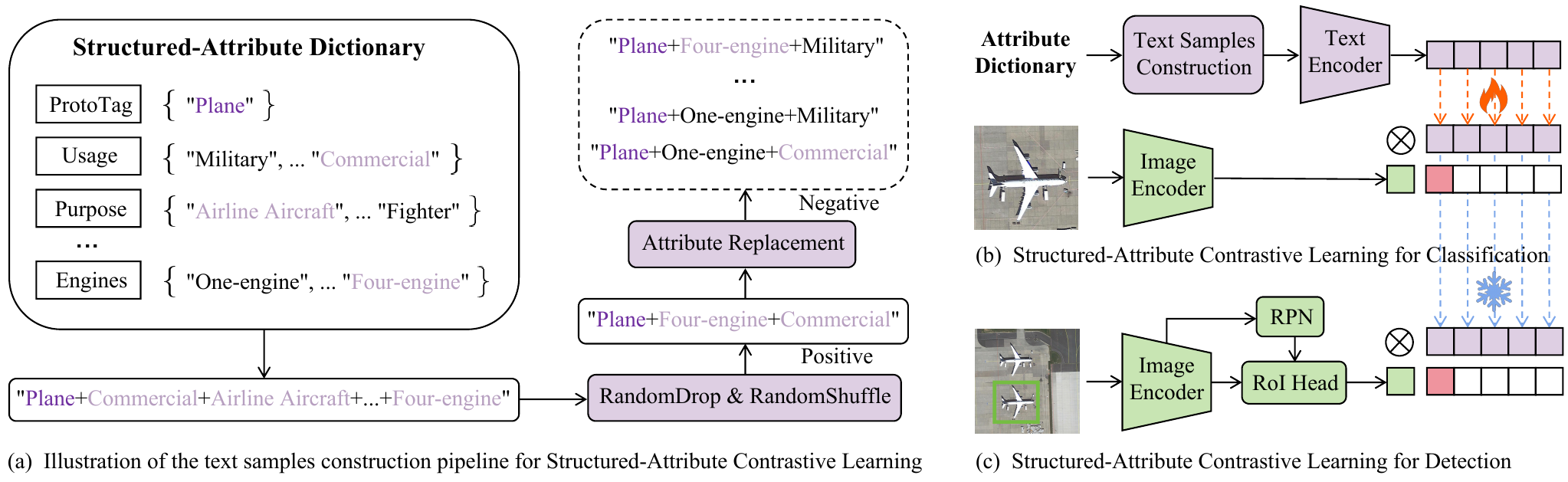} 
    \caption{Overview of Structured-Attribute Contrastive Learning. (a) Positive prompts are generated via Random Drop and Shuffle to enforce permutation invariance. Hard negatives are synthesized via Attribute Replacement, creating counterfactual prompts for fine-grained discrimination. (b) \& (c) Unified Learning: The model optimizes a unified contrastive objective, which aligns representations at the (b) global image level for classification and adapts to the (c) local region level (via RoI features) for detection.
    }
    \label{figure2}
\end{figure*}

\section{Method}

In this section, we present the proposed \method{} framework. 
We begin by revisiting standard contrastive learning, identifying its monolithic supervision relying on data-drive. 
To bridge this gap, we introduce Structured-Attribute Contrastive Learning (SACL). 
As illustrated in Figure~\ref{figure2}, this module initiates a paradigm shift by eformulating contrastive learning as the alignment between visual features and attribute set to enforce fine-grained discriminability. 
Finally, to address the scarcity of required grounding-level attribute data, we propose the Conformal Attribute Reliability Engine (CARE), a self-improving mechanism that distills high-fidelity pseudo-labels to fuel the \method{} training. 

\subsection{Preliminaries}
We first revisit the standard Contrastive Vision-Language Pre-training paradigm. 
Current mainstream methods~\cite{radford2021learning,li2022grounded} typically adopt a Monolithic Alignment strategy. 
Let $f_{I}(\cdot)$ and $f_{T}(\cdot)$ denote the image and text encoders, respectively.
Formally, given a batch of $N$ vision-text pairs $\{(I_i,T_i)\}_{i=1}^N$, where $I_i$ is the visual input and $T_i$ is the corresponding holistic text description (e.g., ~\textit{A satellite view of an aircraft}), the model align visual embedding and text embedding via the InfoNCE loss: 
$$
    \mathcal{L}=-\frac{1}{N} \sum_{i=1}^{N} \log \frac{\exp \left(\operatorname{sim}\left(f_{I}\left(I_{i}\right), f_{T}\left(T_{i}\right)\right) / \tau\right)}{\sum_{j=1}^{N} \exp \left(\operatorname{sim}\left(f_{I}\left(I_{i}\right), f_{T}\left(T_{j}\right)\right) / \tau\right)},
$$
where $sim(\mathbf{u}, \mathbf{v}) = \mathbf{u}^T \mathbf{v} / \| \mathbf{u} \| \| \mathbf{v} \|$ denotes the cosine similarity between normalized $\mathbf{u}$ and $\mathbf{v}$, and $\tau$ is a learnable temperature parameter. 
While effective in general vision, this paradigm relies heavily on brute-force data scaling to compensate for the inherent looseness of In-Batch Negatives. 
In a standard mini-batch, negative texts $\{ T_j \}_{j \ne i}$ typically represent distinct categories (e.g., \textit{ship} vs. \textit{aircraft}). 
These easy negatives allow the model to minimize loss by learning only coarse-grained patterns while ignoring internal structures (e.g., \textit{engine count}). 
In the general domain, this is mitigated by training on billions of pairs to implicitly cover fine-grained descriptions. However, the remote sensing domain confronts a data-driven scalability bottleneck since it lacks the massive fine-grained data required to drive such implicit learning. 
Consequently, relying on this Monolithic Label Learning inevitably induces Semantic Binding by treating categories as inextricable semantic black dots, which results in a recognition blind spot where fine-grained categories collapse into indistinguishable clusters. 

\subsection{Structured-Attribute Contrastive Learning} 
To dismantle the barriers of Monolithic Label Learning, we establish a Structured-Attribute Decoupling Paradigm. 
Instead of mapping visual features to an infinite category space $\mathcal{C}$ ($ |\mathcal{C}| \to \infty$), we decompose objects into orthogonal semantic primitives within a finite structured-attribute space $\mathcal{A}$. 
This formulation compels the model to learn fine-grained representations via combinatorial reasoning rather than mechanical memorization. 

\textbf{Structured Attribute Dictionary Construction.}
We formally redefine the object semantic space. 
Unlike free-form textual descriptions, we postulate that object semantics can be decomposed into a set of structured attributes. 
For an object category $c$, we define its structured attribute dictionary as 
$
\mathcal{D}(c)=\left\{\left(K_{m}, \mathcal{V}_{m}\right)\right\}_{m=1}^{M}. 
$
Here, $K_m$ represents a specific Attribute Dimension (e.g., \textit{engine count}), and $\mathcal{V}_{m}=\left\{v_{m, 1}, v_{m, 2}, \ldots, v_{m, N_{m}}\right\}$ denotes the set of valid Semantic Primitives (discrete states) for this dimension. 
This construction adheres to two physical constraints: 
(1) Dimensional Orthogonality: each key $K_m$ represents an independent axis of visual variation. 
The semantic state of one dimension (e.g., \textit{engine count}) imposes no conditional constraints on another (e.g., \textit{propulsion type}). 
This prior compels the model to learn physically decoupled feature subspaces rather than entangled correlations. 
(2) State Exclusivity: the primitives within a set $\mathcal{V}_{m}$ are mutually exclusive. For any object, exactly one primitive $v_m \in \mathcal{V}_{m}$ holds true (e.g., a plane cannot be simultaneously \textit{four-engine} and \textit{twin-engine}). 
This eliminates label ambiguity and sharpens the decision boundaries. 
Based on this dictionary, we define the ground-truth representation of an object instance $x \in c$ not as a monolithic label, but as a structured attribute set $\mathcal{A}(x)=\{v_{1},v_{2},\ldots,v_{M}\}$, where each $v_m$ is a distinct primitive selected from the dictionary $\mathcal{D}(c)$. 

\textbf{Positive Attribute Prompts Construction.}
We design the positive prompt generation to rigorously model the intrinsic topological structure of the attribute set. Fundamentally, $\mathcal{A}(x)$ is an unordered collection, whereas text encoders are sequence-dependent.
To bridge this gap, we transform the static set into Stochastic Semantic Views via two operations:
(1) Random Shuffling: by applying a random permutation operator $\pi$, we force the model to approximate a permutation-invariant set function, decoupling semantics from spurious positional biases. 
(2) Random Dropout: we randomly mask attributes to generate subsets $\tilde{\mathcal{A}}^{+} \subseteq \mathcal{A}(x)$. 
Mathematically, this approximates sampling from the attribute power set, compelling the model to learn holographic representations where any valid attribute subset aligns with the visual features. Formally, the prompt is generated as 
$
T^+ = \Phi ( \pi ( \tilde{\mathcal{A}}^{+} ) ),
$
where $\Phi$ denotes the serialization function. 
Beyond theory, this theoretical formulation also serves as robust augmentation, preventing overfitting to fixed linguistic templates and simulating partial observability (e.g., occlusion) in remote sensing imagery. 

\textbf{Negative Attribute Prompts Construction.}
Complementing the robustness of positive views, we introduce Counterfactual Hard Negative Mining to strictly enforce fine-grained discriminability. 
Standard in-batch negatives typically represent distinct categories (e.g., \textit{ship} vs. \textit{aircraft}), which are semantically distant from the anchor. 
To dismantle this Semantic Binding, we generate negatives that are topologically adjacent to the positive anchor yet semantically contradictory. 
We execute this via Attribute Replacement. 
Leveraging the State Exclusivity of $\mathcal{D}(c)$, we perturb the positive set $\tilde{\mathcal{A}}^{+}$ by swapping a valid primitive $v_m$ with a hard antagonist $v'_m$ sampled from the same dimension: 
$$
\tilde{\mathcal{A}}^-=(\tilde{\mathcal{A}}^+\setminus\{v_m\})\cup\{v_m^{\prime}\},\quad v_m^{\prime}\in\mathcal{V}_m\setminus\{v_m\}.
$$
The resulting prompt $T^- = \Phi ( \tilde{\mathcal{A}}^{-} )$ acts as a near-manifold hard negative, compelling the model to verify local primitives and strictly refine the decision boundary. 

\textbf{Implementation Details.} 
Unlike standard paradigms that share negatives across a batch, we explicitly construct a unique set of a stochastic positive prompt $T^+$ and $k$ counterfactual hard negatives $\{T^-_{1}, \cdots , T^-_{k}\}$ for each visual instance $I$, while $I$ is the global [CLS] token for classification(figure~\ref{figure2}(b)) and RoI feature for detection(figure~\ref{figure2}(c)). 
For instances lacking attribute annotations(e.g., \textit{Bridge}), we utilize other distinct categories in the dataset as negative samples(e.g., \textit{Intersection} and \textit{Helipad}). 
To ensure stable semantic grounding, we prepend a fixed \textit{ProtoTag} (denoting the coarse category, e.g., \textit{Plane}) to the beginning of every prompt(e.g., \textit{Plane + four engines + swept wings}). 
Crucially, this tag is exempt from the random dropout and shuffling applied to the subsequent attributes. 
The serialization function $\Phi$ adopts a minimalist syntax: it simply concatenates the selected attribute primitives with a ``$+$'' separator (e.g., \textit{delta wing + twin engine}), as shown in Figure~\ref{figure2}.

\begin{table*}[t]
    \centering
    \small
    \setlength{\abovecaptionskip}{2pt}
    \caption{Statistics of the RS-Attribute-15M. Note that `Instances' denotes the bounding box annotations, while `Attributes' refers to the subset of high-fidelity instances annotated with attributes via the CARE engine. RS-O and RS-C denote data without attribute annotation. RS-Attribute-O and RS-Attribute-C are subsets of RS-O and RS-C, respectively. }
    \label{tab1}
    \begin{tabular*}{\textwidth}{@{\extracolsep{\fill}} l c c c c}
        \toprule
        RS-Attribute-15M
        & Source
        & Images
        & Instances
        & Attributes \\
        \midrule
        \multirow{4}{*}{RS-Attribute-O \& RS-O}
        & \multirow{4}{*}{%
            {\tiny
            \begin{tabular}[c]{@{}c@{}}
            AITOD-v2~\cite{xu2022detecting}; RSD-GOD~\cite{zhuang2019single}; DOTA2~\cite{xia2018dota}; \\
            FGSD~\cite{chen2020fgsd}; HRPlane~\cite{bakirman2023benchmark}; DIOR~\cite{li2020object}; \\
            MAR20; SODA-A~\cite{pisani2024soda}; RSOD~\cite{long2017accurate}; \\
            SIMD~\cite{haroon2020multisized}; Fair1M~\cite{sun2022fair1m}; GLHBridge~\cite{li2024learning}; \\
            DroneVehicle~\cite{sun2022drone}; NWPU-VHR-10~\cite{cheng2014multi};
            \end{tabular}}
        }
        & \multirow{4}{*}{216,895}
        & \multirow{4}{*}{3,058,905}
        & \multirow{4}{*}{881,672} \\
        & & & & \\
        & & & & \\
        & & & & \\
        \midrule
        RS-Attribute-C \& RS-C
        & Diverse third-party channels
        & 570,853
        & 28,092,216
        & 14,609,802 \\
        \bottomrule
    \end{tabular*}
\end{table*}

\subsection{Conformal Attribute Reliability Engine}
The efficacy of \method{} hinges on the availability of grounding-level attribute-image data, which is scarce in existing datasets. To bridge this gap, we curate an attribute dataset covering the three most ubiquitous remote sensing categories(\textit{plane, ship, vehicle}) via the proposed Conformal Attribute Reliability Engine (CARE). 
CARE operates via a progressive Tune-Calibrate-Scale pipeline, designed to rigorously distill high-fidelity supervision from noisy sources with statistical guarantees. 

\textbf{Tune Stage.}
The primary objective of this stage is to establish a specialized attribute classifier capable of generating high-quality instance-level pseudo-labels for subsequent scaling. 
We first address the cold-start problem by curating an expert-verified seed attribute classification dataset $\mathcal{D}_{seed}$. 
Acknowledging the varying resolutions and structural complexities of remote sensing categories, we employ an adaptive granularity strategy for attribute definition: 
1) For large, structurally discernable targets (e.g., \textit{planes}), we define component-level attributes (e.g., \textit{wing shape}, \textit{engine position}) to capture rich details; 
2) For smaller targets (e.g., \textit{ships}, \textit{vehicles}), where component visibility is limited, we focus on functional-level attributes (e.g., \textit{container ship}, \textit{pick-up}). 
More detailed attribute definitions are provided in Table~\ref{tab7}. 
Upon this foundation, we fine-tune a teacher model (initialized from RemoteCLIP-ViT/32B~\cite{liu2024remoteclip}) using the SACL strategy. 
This process yields RemoteCLIP-FG, which serves two critical purposes: it equips the teacher with fine-grained discrimination capabilities needed for pseudo-labeling, and crucially, it aligns the text encoder with visual concepts, providing a robust initialization for the subsequent detector training. 

\textbf{Calibrate Stage.}
Although RemoteCLIP-FG exhibits strong attribute classification capability, its raw confidence distribution is often misaligned with empirical accuracy (e.g., a confidence score of 0.9 does not imply 90\% accuracy). 
Applying a heuristic fixed threshold to these uncalibrated scores introduces unpredictable label noise. 
To ensure high-fidelity supervision, we employ Conformal Prediction~\cite{shafer2008tutorial} to derive statistically valid, attribute-specific thresholds. 
For an attribute class $k$, we reserve a calibration subset $\mathcal{D}_{cal}^k$ and define the non-conformity score function as $s_{k}(x)=1-\hat{p}_{k}(x)$, where $\hat{p}_{k}(x)$ denotes the predicted probability for the ground truth. 
To control the error rate at a predefined tolerance level $\alpha$ (e.g., 0.1), we compute the $(1-\alpha)$-th quantile of the non-conformity scores on $\mathcal{D}_{cal}^k$, denoted as $\hat{q}_{k}$. 
The calibrated threshold is then derived as $\tau_{k}=1-\hat{q}_{k}$. 
Theoretically, this mechanism provides a marginal coverage guarantee ensuring that the truth is retained in the prediction set with probability at least $1-\alpha$:
$$
\mathbb{P}(y \in \mathcal{C}(x)) \ge 1 - \alpha, \ where \ \ \mathcal{C}(x) = \{ k \mid \hat{p}_k(x) \ge \tau_k \}. 
$$
By replacing heuristic guesswork with rigorous statistical bounds, this stage purifies the pseudo-labels to reject uncertain predictions while preserving high-confidence attributes. 

\textbf{Scale Stage.}
Equipped with the statistically calibrated thresholds $\tau_{k}$, we execute a large-scale filtration pipeline to construct the final RS-Attribute-15M detection dataset. 
This dataset is aggregated from two distinct sources: 
1) RS-Attribute-O, which revitalizes 14 open-source RS detection datasets and is enriched with fine-grained attribute annotations by the CARE; 
2) RS-Attribute-C, sourced from diverse third-party channels, where we first generate instance-level detection pseudo-labels and subsequently append fine-grained attribute pseudo-labels via the CARE. 
This rigorous filtration yields RS-Attribute-15M (details are shown in Table~\ref{tab1}), the largest attribute-grounded RS dataset to date with over 15 million annotated instances, providing the essential data fuel for learning fine-grained attribute representations.

\begin{table*}[t]
    \centering
    \small
    \setlength{\abovecaptionskip}{2pt}
    \caption{Comparison with existing methods on DOTA-v2.0 (HBB) and DIOR-H. \method{} denotes the baseline trained without attribute supervision, while \method{}* refers to the model trained with our SACL strategy incorporating fine-grained attribute annotations. The base detectors of LAE-DINO and OpenRSD are DINO and RTMDet, respectively.}
    \label{tab2}
    \begin{tabular}{l l l c c c}
        \toprule
        Methods 
        & Backbone 
        & Training data 
        & \multicolumn{2}{c}{DOTA2 (mAP)} 
        & DIOR (AP50) \\
        \cmidrule(lr){4-5}
        & & 
        & Zero-shot 
        & Fine-tune 
        & Fine-tune \\
        \midrule
        Faster-RCNN~\cite{ren2016faster}
        & ConvNeXT-T ~\cite{liu2022convnet}
        & DOTA2.0 \& DIOR 
        & -- 
        & 39.21 
        & 74.10 \\

        DINO~\cite{zhang2022dino}
        & Swin-T ~\cite{liu2021swin}
        & DOTA2.0 \& DIOR 
        & -- 
        & 38.32 
        & 73.27 \\

        RTMDet~\cite{lyu2022rtmdet}
        & RTMDet-L ~\cite{lyu2022rtmdet}
        & DOTA2.0 \& DIOR 
        & -- 
        & 39.43 
        & 76.52 \\

        LAE-DINO~\cite{pan2025locate} 
        & Swin-T 
        & LAE-1M 
        & -- 
        & 39.38 
        & 75.67 \\

        OpenRSD~\cite{huang2025openrsd}
        & RTMDet-L 
        & ORSD+ 
        & -- 
        & 44.57 
        & 76.70 \\

        ViTP~\cite{li2025visual}
        & ViT-L~\cite{dosovitskiy2020image}
        & DIOR 
        & -- 
        & -- 
        & 79.80 \\
        \midrule
        \method{} 
        & ConvNeXT-T 
        & RS-O 
        & -- 
        & 41.24 
        & 76.10 \\

        \method{}
        & ConvNeXT-T 
        & RS-C 
        & 35.63 
        & 42.56 
        & -- \\

        \method{}
        & ConvNeXT-T 
        & RS-O + RS-C 
        & -- 
        & 45.10 
        & 79.97 \\

        \method{}*
        & ConvNeXT-T 
        & RS-Attribute-15M 
        & -- 
        & 44.43 
        & 79.94 \\

        \method{}*
        & ConvNeXT-L 
        & RS-Attribute-15M 
        & -- 
        & 47.14 
        & 83.10 \\
        \bottomrule
    \end{tabular}
\end{table*}

\section{Experiments}

\subsection{Implementation Details}

We implement \method{} based on the Faster R-CNN~\cite{ren2016faster} detector, as its label assignment strategy naturally accommodates ambiguous samples, enabling more stable training under noisy pseudo labels compared with the hard binary matching strategies adopted by one-stage and DETR-based detectors. 
Furthermore, we replace the original MaxIoU label assignment strategy with RFLA~\cite{xu2022rfla} to better accommodate the characteristics of remote sensing objects. 
The visual backbone employs ConvNeXT-T/L~\cite{liu2022convnet} initialized with DINOv3~\cite{simeoni2025dinov3} weights, while the text encoder utilizes the frozen RemoteCLIP-FG to maintain semantic stability. 
All experiments are conducted on 8 NVIDIA A40 GPUs using the AdamW optimizer with a batch size of 8, an initial learning rate of $1\times10^{-5}$, and a weight decay of $5\times10^{-2}$. 
The models are trained for 12 epochs, with the learning rate decayed at the 8th and 11th epochs. 

\subsection{Zero-Shot and Supervised Transfer}
To comprehensively evaluate the transferability of \method{}, we conduct experiments under two distinct settings: zero-shot domain transfer and supervised transfer. 
These evaluations are designed to verify not only the model's cross-domain adaptability but also the generalization efficacy of the proposed paradigm on common categories. 

\textbf{Datasets and Evaluation Details.}
We first employ two benchmarks, DOTA-v2.0 (HBB)~\cite{xia2018dota} and DIOR-H~\cite{li2020object}, to evaluate the model's adaptability to standard scenarios and categories.
For DOTA-v2.0 (HBB), we adhere to the official single-scale evaluation protocol. 
Specifically, high-resolution images are cropped into patches of $1024 \times 1024$ with an overlap of 200 pixels. 
To validate the model's capabilities, we adopt specific training protocols:
(1) for zero-shot domain transfer, the model is trained exclusively on our custom RS-C dataset; 
(2) for supervised transfer, we integrate the training set into the training data. 
We report the performance on the validation set. 
For the DIOR-H benchmark, following established practices~\cite{li2025visual}, we crop images into $800 \times 800$ patches.
We merge the training and validation sets for fine-tuning and report the final performance on the testing set.

\begin{table*}[t]
\centering
\small
\setlength{\abovecaptionskip}{2pt}
\caption{Quantitative results of atomic fine-grained attribute recognition on the constructed benchmark. `P' for Purpose, `U' for Usage, `EP' for Engine Position, `WC' for Wing Configuration, `NE' for Number of Engine, `PT' for Propulsion Type, `S' for Subcategory.}
\label{tab3}
\setlength{\tabcolsep}{3pt}
    \begin{tabular}{l l l c c c c c c c c c c c}
        \toprule
        Methods & Backbone & Training data 
        & \multicolumn{6}{c}{Plane (mAP)} 
        & \multicolumn{3}{c}{Ship (mAP)} 
        & \multicolumn{2}{c}{Vehicle (mAP)} \\
        \cmidrule(lr){4-9}
        \cmidrule(lr){10-12}
        \cmidrule(lr){13-14}
        & & 
        & \makecell{P}
        & \makecell{U}
        & \makecell{EP}
        & \makecell{WC}
        & \makecell{NE}
        & \makecell{PT}
        & \makecell{P}
        & \makecell{U}
        & \makecell{S}
        & \makecell{P}
        & \makecell{U} \\
        \midrule
        LAE-DINO
        & Swin-T
        & LAE-1M
        & 1.04
        & 6.32
        & --
        & --
        & --
        & --
        & 8.41
        & 5.47
        & 9.84
        & 14.74
        & 12.31 \\
        
        OpenRSD
        & RTMDet-L
        & ORSD+
        & 9.03
        & 23.07
        & --
        & --
        & --
        & --
        & 15.41
        & 12.36
        & 15.67
        & 15.53
        & 18.35 \\
        
        \midrule
        \method{}
        & ConvNeXT-T
        & RS-Attribute-O
        & 67.61
        & 48.77
        & 68.47
        & 70.14
        & 74.27
        & 71.80
        & 20.40
        & 34.00
        & 18.33
        & 19.36
        & 24.72 \\
        
        & ConvNeXT-T
        & RS-Attribute-15M
        & 64.85
        & 52.66
        & 71.33
        & 62.60
        & 77.06
        & 76.35
        & 27.26
        & 44.52
        & 17.61
        & 23.90
        & 28.80 \\
        
        & ConvNeXT-L
        & RS-Attribute-15M
        & 72.04
        & 57.17
        & 77.33
        & 69.54
        & 80.47
        & 79.10
        & 27.82
        & 48.03
        & 27.61
        & 23.93
        & 31.35 \\
        \bottomrule
    \end{tabular}
\end{table*}

\begin{table}[t]
\centering
\small
\setlength{\abovecaptionskip}{2pt}
\caption{Evaluation of compositional attribute recognition(mAP). Rows 1 and 2 correspond to \method{}-T and \method{}-L. Columns indicate the number of combined attributes, reporting the average performance across all valid combinations.}
\label{tab4}
    \begin{tabular}{cccccccc}
        \toprule
        \multicolumn{5}{c}{Plane} & \multicolumn{2}{c}{Ship} & Vehicle \\
        \cmidrule(lr){1-5} \cmidrule(lr){6-7} \cmidrule(lr){8-8}
        2 & 3 & 4 & 5 & 6 & 2 & 3 & 2 \\
        \midrule
        62.1 & 66.9 & 66.7 & 60.6 & 58.8 & 22.5 & 15.6 & 24.0 \\
        69.2 & 72.3 & 74.4 & 72.6 & 72.1 & 25.2 & 25.4 & 23.1 \\
        \bottomrule
    \end{tabular}
    \vspace{-10pt}
\end{table}

\textbf{Main Results.}
The quantitative results in Table~\ref{tab2} empirically substantiate both the robust cross-domain generalization and the superior representation quality on common categories delivered by our paradigm. 
In the zero-shot setting, training exclusively on the custom RS-C subset yields a remarkable 35.63\% mAP on DOTA-v2.0. 
Regarding supervised transfer, our lightweight ConvNeXT-T model achieves 45.10\% on DOTA-v2.0 and 79.97\% on DIOR, significantly surpassing the specialized OpenRSD~\cite{huang2025openrsd} (44.57\% and 76.70\%). 
These underscore the intrinsic robustness of our data, validating its ability to cover diverse semantic scenarios even without target-domain supervision. 
Scaling to ConvNeXT-L further elevates performance to 47.14\% and 83.10\%, demonstrating the substantial scalability of our \method{}. 
Finally,  we address the marginal performance fluctuation (e.g., 45.10\% vs. 44.43\%) observed when incorporating fine-grained attribute supervision. 
This slight trade-off is an expected consequence of the optimization shift that the model diverts capacity from overfitting coarse category boundaries to learning intricate internal structures. 
Crucially, this negligible cost is well-justified, as it strategically refines the representation space, enabling the coverage of a much broader spectrum of fine-grained categories as demonstrated in subsequent sections. 

\subsection{Fine-grained and Attribute Recognition}

We then evaluate the fine-grained and attribute recognition capabilities of \method{}. 
Our objective is to verify that \method{} empowers the model to not only identify visual attributes but also leverage them to expand recognition granularity. 

\textbf{Benchmark and Evaluation Details.}
To evaluate fine-grained understanding capabilities, we construct a specialized benchmark derived from the authoritative datasets, including Fair1M~\cite{sun2022fair1m}, MAR20~\cite{wenqi2024mar20}, and HRSC2016. 
This benchmark encompasses three categories (plane, ship, and vehicle) and is meticulously annotated with hierarchical attributes, ranging from high-level functional descriptions (e.g., \textit{Purpose}) to physical components (e.g., \textit{Engine position}). 
The details are shown in Table~\ref{tab9}. 
We adopt a strict prompt-matching protocol to verify the model's response to specific attributes under two settings:
(1) Atomic Attribute Verification: we construct prompts by concatenating the category name with a specific attribute value (formatted as $\{Category\}+\{Attribute Value\}$, e.g., \textit{`Plane+Eight-engine'}) to test the model's sensitivity to individual attribute. 
(2) Compositional Attribute Verification: to further assess multi-attribute recognition robustness, we generate complex prompts by permuting and combining values from different attribute keys (e.g., combining \textit{Wing Configuration} with \textit{Engine Position}). 
This setting verifying models' capability to expand the spectrum of recognizable categories through attribute composition. 
For quantitative evaluation, we report the average performance across all combinations with the same number of attributes. 

\begin{figure}[t]
  \centering
  \setlength{\abovecaptionskip}{0.cm}
  \includegraphics[width=\linewidth]{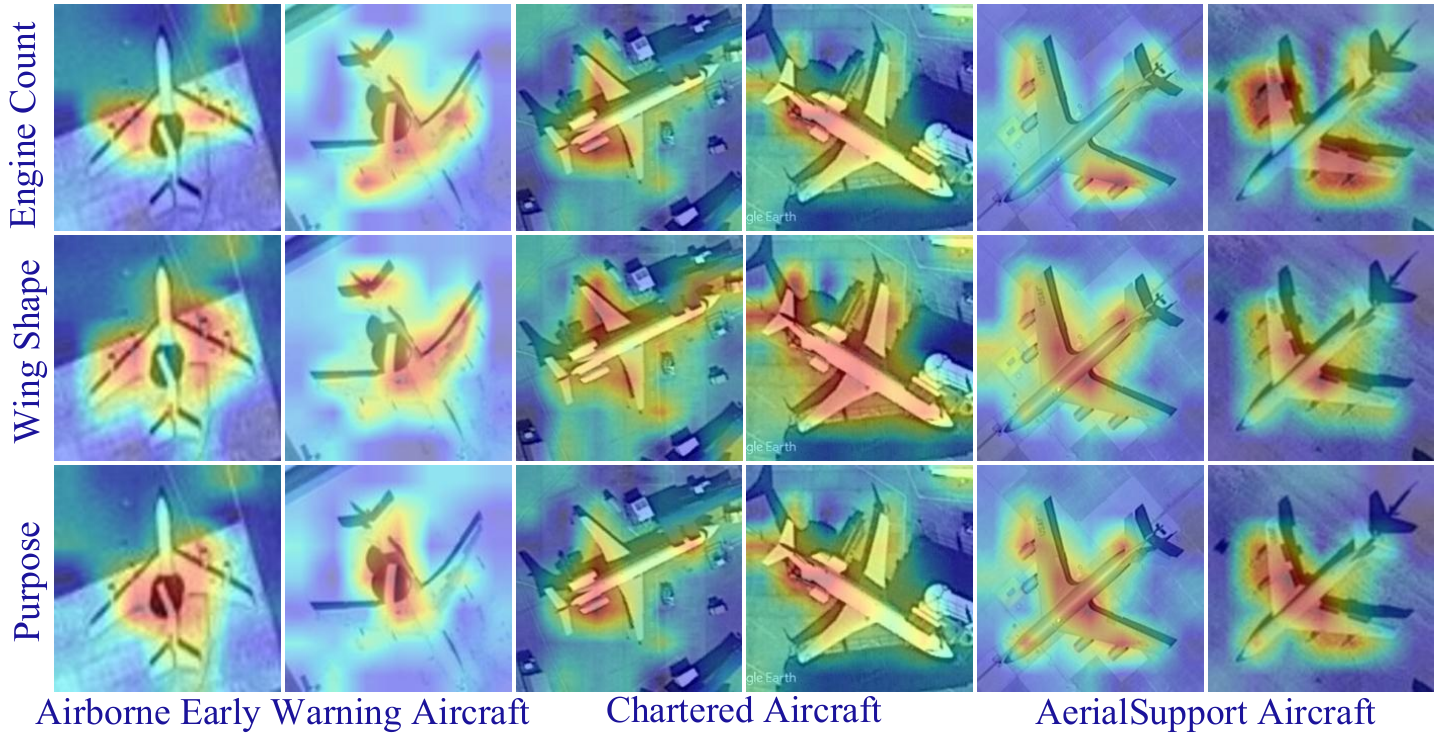}
  \caption{Visualizations of \method{} for attribute-guided fine-grained detection. \method{} not only identifies fine-grained categories but also enables recognition based on specific attributes. }
  \label{figure3}
  \vspace{-10pt}
\end{figure}

\begin{figure*}[t] %
    \footnotesize
    \setlength{\abovecaptionskip}{0pt}
    \centering
    \includegraphics[width=\linewidth]{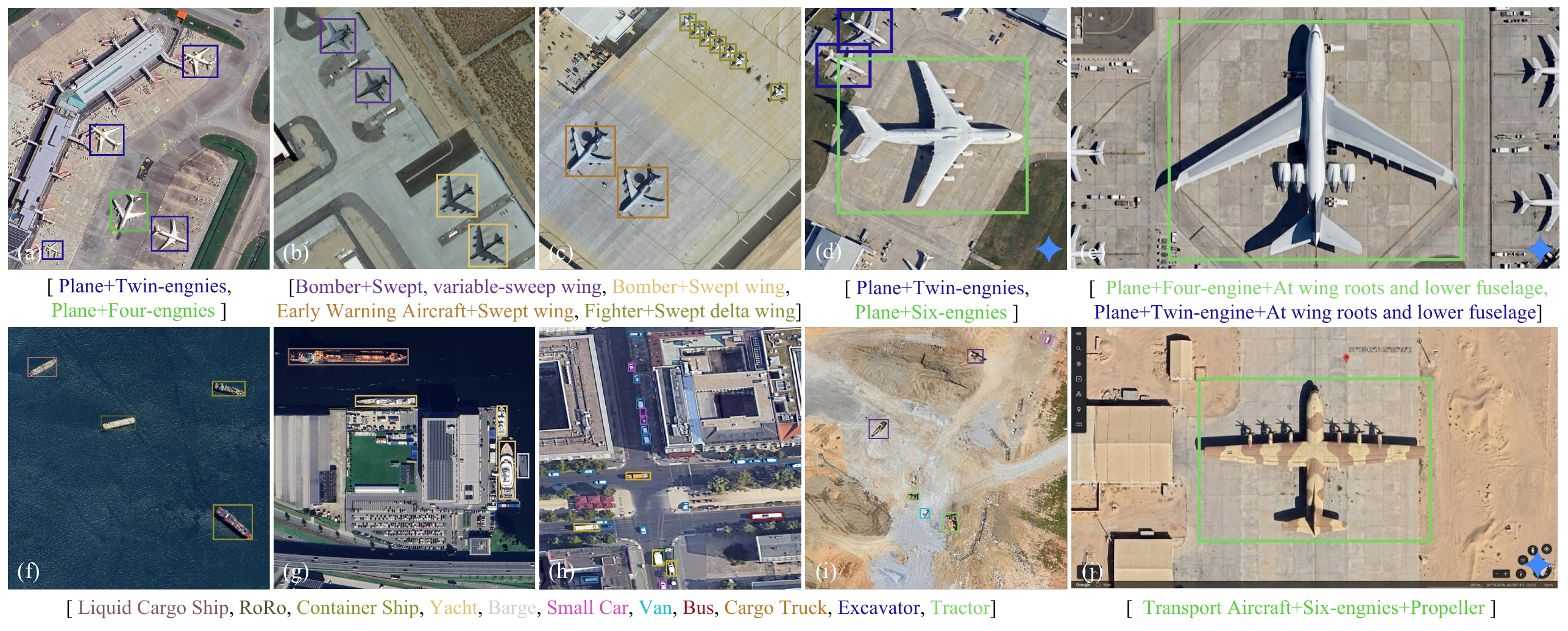} 
    \caption{Visualization of detection results using SLIP-RS. The figure showcases the model's ability to localize targets based on compositional attributes and identify fine-grained categories. Note that the samples in (d), (e), and (j) are generated by Nano Banana Pro.
    }
    \label{figure4}
\end{figure*}

\textbf{Main Results.} 
Table~\ref{tab3} reports the quantitative comparison on atomic attribute verification. 
Existing methods struggle to distinguish fine-grained classes and attributes. 
For instance, on \textit{Plane Purpose}, OpenRSD achieves only 9.03\% mAP, failing to differentiate functional types like bombers and fighters. 
In contrast, \method{} (ConvNeXT-L) achieves 72.04\%, representing a significant improvement. 
This failure stems from their reliance on brute-force category enumeration, thus are constrained by the scarcity of fine-grained remote sensing data. 
Finally, we assess the potential to expand recognizable categories through attribute composition. 
Instead of learning fixed vocabularies, we evaluate the model's ability to localize targets defined by arbitrary combinations of learned attributes. 
As shown in Table~\ref{tab4}, \method{} achieves robust performance on multi-attribute prompts.
This capability is visually corroborated in Figure~\ref{figure4}. 
The model not only precisely distinguishes fine-grained categories but also successfully localizes targets defined by attribute permutations, effectively validating the feasibility of category expansion through attribute composition. 

\textbf{Qualitative Analysis and Visualization.}
To interpret the learned representations, we visualize activation maps for different attribute keys (Figure~\ref{figure3}). 
The results demonstrate that \method{} precisely grounds semantic concepts into visual primitives.
For instance, in the rows for \textit{Engine Count}, the high-response regions are strictly concentrated on the engines, ignoring other fuselage parts. 
Similarly, for \textit{Wing Shape} queries, the attention map specifically highlights the wing boundaries, verifying that the model explicitly reasons about geometry rather than global context.
Crucially, for abstract \textit{Purpose} queries, the model identifies key functional discriminators. 
For instance, it explicitly highlights the circular rotodome for \textit{Airborne Early Warning Aircraft} and the refueling probe for \textit{AerialSupport Aircraft}. 
This confirms that the model reasons based on physically meaningful features rather than spurious background correlations. 
We also utilize NanoBanana Pro to synthesize counterfactual samples featuring attribute configurations unseen during training. 
As illustrated in Figure~\ref{figure4}, the model successfully identifies aircraft with \textit{Six Engines} (an attribute cardinality absent in the training set) and correctly localizes targets with unconventional layouts, such as \textit{Plane+Four-engine+At wing roots and lower fuselage}, which suggests that the model may capture attributes as partially decoupled visual cues rather than relying solely on holistic label matching. 
While a systematic evaluation of zero-shot attribute composition is beyond the scope of this study, we believe these preliminary observations highlight an avenue for future research in open-world perception. 

\subsection{Evaluation on Complex Downstream Tasks}
We further extend our evaluation to two challenging downstream scenarios to validate the robustness of \method{} as a pre-training method. 

\begin{table}[t]
    \centering
    \small
    \setlength{\abovecaptionskip}{2pt}
    \caption{Comparison of different methods on the Xview dataset.}
    \label{tab5}
    \begin{tabular}{l l c}
        \toprule
        Methods & Backbone & Xview (AP50) \\
        \midrule
        MTP~\cite{wang2024mtp}       & ViT-L         & 21.5 \\
        ViTP~\cite{li2025visual}     & ViT-L         & 26.6 \\
        \method{}                     & ConvNeXT-T    & 25.9 \\
        \method{}*                    & ConvNeXT-T    & 27.2 \\
        \bottomrule
    \end{tabular}
\end{table}

\begin{table}[t]
    \centering
    \small
    \setlength{\abovecaptionskip}{2pt}
    \caption{Evaluation on the DOTAv-2.0 (OBB) dataset.}
    \label{tab6}
    \begin{tabular}{l l l c}
        \toprule
        Methods & Backbone  & Params & mAP \\
        \midrule
        BFM~\cite{cha2023billion}         & ViT-G       & 2.4B         & 58.69 \\
        S5~\cite{lv2025s5}                & ViT-H       & 671.7M      & 59.89 \\
        ViTP~\cite{li2025visual}          & ViT-L       &  347.9M     & 60.23 \\
        \method{}                         &  ConvNeXT-L & 198M        & 62.54 \\
        \bottomrule
    \end{tabular}
    \vspace{-10pt}
\end{table}

\textbf{Transfer to Large-scale Fine-grained Recognition.}
We first investigate whether the intrinsic fine-grained representations established via attribute supervision can effectively transfer to downstream fine-grained recognition task.
The underlying hypothesis is that alignment with semantic attributes induces a highly discriminative feature manifold, which serves as a superior initialization for resolving subtle inter-class discrepancies during subsequent fine-tuning. 
To verify this, we fine-tune \method{} on Xview dataset, a challenging benchmark encompassing 60 fine-grained categories. 
Specifically, following prior works~\cite{wang2024mtp}, we randomly select 700 images for training and 146 images for testing. 
The images are cropped into $800 \times 800$ patches with an overlap of 200 pixels. 
Quantitative results are presented in Table~\ref{tab5}.
Incorporating structured-attribute contrastive learning (SLIP-RS*) boosts the baseline AP50 from 25.9\% to 27.2\%, outperforming even the ViT-L based ViTP (26.6\%) despite using a lightweight ConvNeXt-T. 
This improvement confirms that attribute supervision forces the model to encode critical part-level discriminative cues, which are indispensable for resolving subtle inter-class ambiguities in dense semantic spaces. 

\textbf{Oriented Object Detection.}
To further assess the generalization ability of the learned representations, we extend our evaluation to the DOTA-v2.0 (OBB) benchmark.
Specifically, we extract the pre-trained ConvNeXT-L backbone from \method{} and integrate it with the Oriented R-CNN detector.
We crop images into $1024 \times 1024$ with an overlap of 200 pixels. 
Following standard protocols, we merge the training and validation sets for training and report the performance on the test set. 
Finally, Table~\ref{tab6} shows \method{} achieving 62.54\% mAP. 
This result rivals recent billion-scale foundation models, confirming its exceptional capability as a robust foundation for adaptation to complex downstream scenarios. 

\section{Conclusions} 
We pioneer a Structured-Attribute Decoupling Paradigm to transcend monolithic learning bottlenecks. 
Leveraging SACL and the CARE,  \method{} unlocks robust fine-grained representations without exhaustive supervision. 
Experiments confirm its unprecedented performance, establishing a scalable foundation for remote sensing object detection. 

\section*{Acknowledgments}

This research was supported by the Fund of the National Natural Science Foundation of China (62522607, 62495061, 62276145 and 62576177), Shenzhen Science and Technology Program (QNXMB20250701090801002, JCYJ20250604184027034, JCYJ20240813114237048), Guangdong Basic and Applied Basic Research Foundation (2026A1515011435), and the Fundamental Research Funds for the Central Universities (Nankai University). 

\section*{Impact Statement}

This paper presents work whose goal is to advance the field of Machine
Learning. There are many potential societal consequences of our work, none
which we feel must be specifically highlighted here. 

\nocite{langley00}

\bibliography{example_paper}
\bibliographystyle{icml2026}

\newpage
\appendix
\onecolumn

\section{Implementation Details of Structured-Attribute Contrastive Learning}

Unlike standard paradigms that utilize batch-shared negatives, we implement an instance-specific contrastive formulation to enforce fine-grained discrimination. 
For each visual instance $I$ that represented as the global [CLS] token for classification (Figure~\ref{figure2}(b)) or RoI features for detection (Figure~\ref{figure2}(c)), we explicitly construct a unique candidate set $\mathcal{T}_I = \{T^+, T^-_1, \dots, T^-_k\}$. 
The positive prompt $T^+$ is generated via a serialization function $\Phi$ that adopts a minimalist syntax, concatenating a fixed \textit{ProtoTag} (denoting the coarse category, e.g., \textit{Plane}) with selected attribute primitives using a ``$+$'' separator (e.g., \textit{Plane + delta wing + twin engine}). 
To compel the model to comprehend intrinsic visual logic, we synthesize $k$ counterfactual hard negatives $\{T^-_i\}$ by strategically corrupting the semantic structure of $T^+$. 
Specifically, we randomly replace attributes with contradictory primitives from the same dimension (e.g., swapping \textit{twin-engine} for \textit{four-engine}), while strictly keeping the \textit{ProtoTag} invariant to ensure stable semantic grounding. 
For instances lacking attribute annotations, we utilize other distinct categories in the dataset as negative samples. 
Taking the DOTA-v2.0 dataset as an example, if a \textit{Plane} instance lacks specific attribute labels, its negative set $\{T^-_i\}$ is constructed by sampling from the other 17 distinct categories (e.g., \textit{Ship}, \textit{Bridge}, \textit{Harbor}, ...), ensuring the model maintains robust categorical boundaries even in the absence of granular supervision. 
The contrastive loss is then computed exclusively within this instance-wise set $\mathcal{T}_I$ : 
$$
\mathcal{L}_{\mathrm{SACL}}
= - \log
\frac{ \exp\left( sim(f_{I}(I_i), f_{T}(T^+_i))  / \tau \right)}
{ \exp\left( sim(f_{I}(I_i), f_{T}(T^+_i)) / \tau \right) + \sum_{j}^{k}
\exp\left( sim(f_{I}(I_i), f_{T}(T^-_j)) / \tau \right)}, 
$$
where $sim(\mathbf{u}, \mathbf{v}) = \mathbf{u}^T \mathbf{v} / \| \mathbf{u} \| \| \mathbf{v} \|$ denotes the cosine similarity between normalized $\mathbf{u}$ and $\mathbf{v}$, and $\tau$ is a learnable temperature parameter. 
It optimizes the alignment between visual features and their precise attribute configurations while suppressing highly plausible but factually incorrect descriptions. 

\section{Implementation Details of Conformal Attribute Reliability Engine} 

\begin{table*}[h]
    \centering
    \small
    \caption{Details of the seed attribute classification dataset with attribute statistics for each category.}
    \label{tab7}
    \begin{tabular*}{\textwidth}{@{\extracolsep{\fill}} l l}
        \toprule
        Category & Details(The numbers represent the quantity) \\
        \midrule
        \multirow{4}{*}{Plane}
        & \multirow{4}{*}{%
            {\tiny
            \begin{tabular}[l]{@{}l@{}}
                \textbf{Number of Images:}: 
                29951. \\
                \textbf{ProtoTag}: 
                Plane (29951). 
                \textbf{Propulsion type}: 
                Jet (25586);
                Propeller (4365). 
                \textbf{Number of engines}: 
                One-engine (2543); 
                Eight-engine (957);
                Four-engine (11100);
                Twin-engine (15351).  \\
                \textbf{Engine position}: 
                One the nose (853); 
                At wing roots and lower fuselage (2758);
                Beneath the wings (16121); 
                Embedded within wing (276); 
                Above the wings (2); 
                Rear fuselage (9941). \\ 
                \textbf{Purpose}: 
                Aerial Support Aircraft (2591); \\
                Airborne Early Warning Aircraft (691); 
                Airline Aircraft (7747); 
                Anti-Submarine Warfare Aircraft (1468);
                Bomber (3352); 
                Trainer (1035); 
                Chartered aircraft (676);
                Attack aircraft (203);  \\
                Fighter (9718);
                Transport Aircraft (2289).
                \textbf{Usage}: 
                Civilian Aircraft (1856);
                Commercial Aircraft (7747);
                Military Aircraft (20348). 
                \textbf{Wing configuration}: 
                Straight wing (3172); \\
                Swept diamond-like wing (409); 
                Flying wing (276); 
                Swept delta wing (185); 
                Swept wing (23835);
                Swept, variable-sweep wing (2074).
            \end{tabular}}
        } \\
        &  \\
        &  \\
        &  \\
        &  \\
        \midrule
        \multirow{2}{*}{Ship}
        & \multirow{3}{*}{%
            {\tiny
            \begin{tabular}[l]{@{}l@{}}
                \textbf{Number of Images:}: 
                10265. \\
                \textbf{ProtoTag}: 
                Ship (10265). 
                \textbf{Purpose}: 
                Aircraft Carrier (238);
                Amphibious Ship (389);
                Auxiliary Ship (118);
                Cargo Ship (2686);
                Commander (75);
                Cruiser (258); \\
                Destroyer (698);
                Fishing Vessel (1079);
                Frigate (562);
                Medical Ship (4);
                Military Transport Ship (164);
                Motorboat (496); 
                Passenger Ship (823);
                Patrol (60); \\
                Submarine (550);
                Landing (53); 
                Medical Ship (13); 
                Tugboat (1257). 
                \textbf{Subcat}: 
                Barge (11);
                Container Ship (51);
                Cruise Ship (47);
                Dry Cargo Ship (3843);
                Liquid Cargo Ship (597); \\
                RoRo (68); 
                Yacht (416). 
                \textbf{Usage}: 
                Civilian Ship (2442);
                Commercial Ship (2686);
                Engineering Ship (1871);
                Military Ship (3266).
            \end{tabular}}
        } \\
        &  \\
        &  \\
        &  \\
        \midrule
        \multirow{1}{*}{Vehicle}
        & \multirow{1}{*}{%
            {\tiny
            \begin{tabular}[l]{@{}l@{}}
                \textbf{Number of Images:}: 
                41893. \\
                \textbf{ProtoTag}: 
                Vehicle (41893). 
                \textbf{Purpose}: 
                Bus (1063);
                Cargo Truck (6011);
                Dump Truck (585);
                Excavator (1544);
                Pick-up (859);
                Small Passenger Car (20855); \\
                Tractor (585); 
                Truck Tractor (2024);
                Van (7610). 
                \textbf{Usage}: 
                Engineering Vehicle (2886);
                Large Civilian Vehicle (1063);
                Small Civilian Vehicle (29324);
                Truck (8620).
            \end{tabular}}
        } \\
        &  \\
        \bottomrule
    \end{tabular*}
\end{table*}

\subsection{Tune Stage.} 
\begin{figure*}[ht] %
    \footnotesize
    \setlength{\abovecaptionskip}{0pt}
    \centering
    \includegraphics[width=\linewidth]{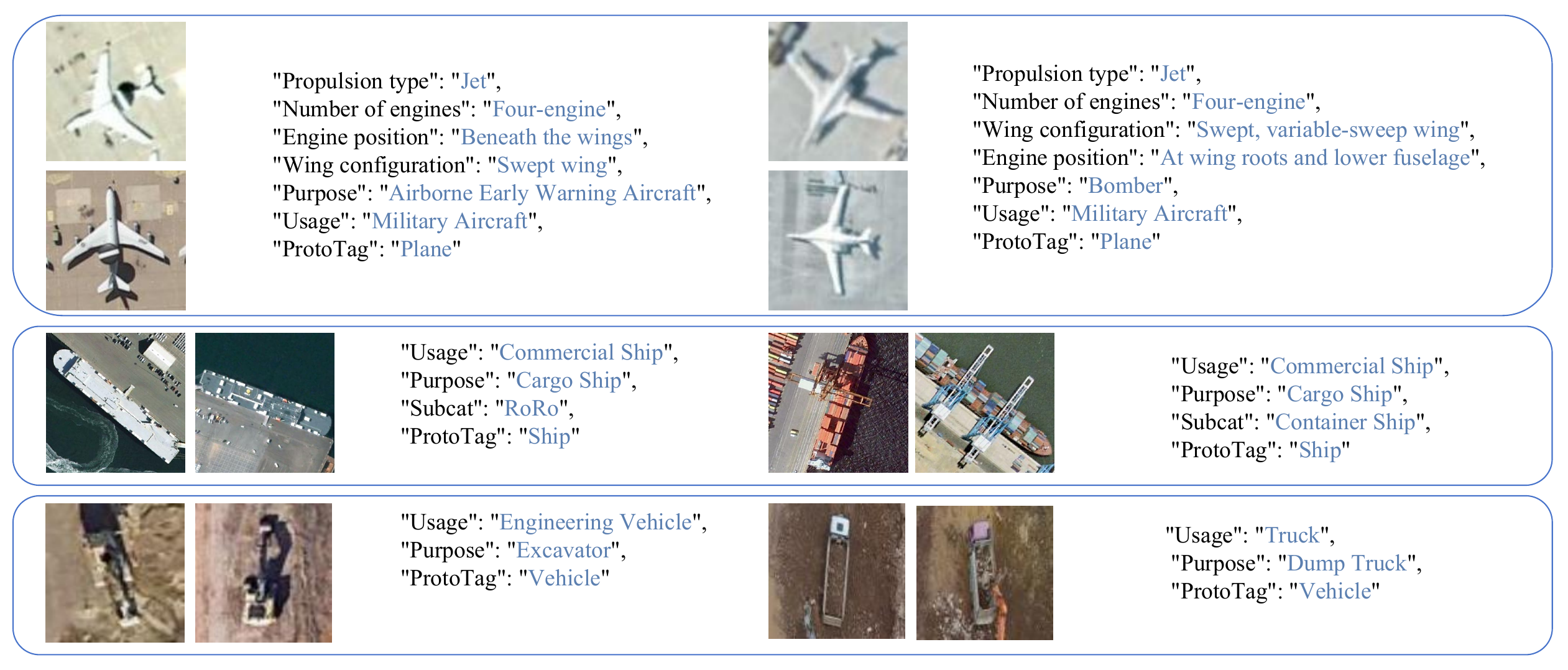} 
    \caption{Demo examples from the Seed Attribute Classification Dataset.}
    \label{figure5}
\end{figure*}

To initiate the pipeline, we construct a high-quality Seed Attribute Classification Dataset, focusing on three of the most ubiquitous yet challenging categories in remote sensing (\textit{Plane, Vehicle, Ship}). 
Detailed statistics of this seed corpus are provided in Table~\ref{tab7}, with representative visual exemplars illustrated in Figure~\ref{figure5}. 
Leveraging this dataset in conjunction with our proposed SACL strategy, we fine-tune the RemoteCLIP-ViT/32B foundation model to serve as a specialized fine-grained teacher (termed as RemoteCLIP-FG), which is subsequently deployed to generate attribute pseudo-labels for every instance across the object detection datasets. 
We implement a comprehensive adaptation strategy by injecting Low-Rank Adaptation (LoRA) modules into every layer of both the Vision Transformer (ViT) and the Text Encoder. 
The LoRA rank and alpha scaling factor are set to $r=2$ and $\alpha=1$, respectively. 
RemoteCLIP-FG is optimized using the AdamW optimizer with an initial learning rate of $1 \times 10^{-5}$, a weight decay of $1 \times 10^{-2}$, and $\beta$ coefficients of $(0.9, 0.999)$. 
We train the model for 100 epochs, employing a cosine decay schedule to decay the learning rate to a minimum of $1 \times 10^{-6}$ every epoch. 

\subsection{Calibrate Stage.} 
To ensure the robustness of the conformal thresholds, we construct a diverse calibration subset $\mathcal{D}_{cal}$, with detailed statistics provided in Table~\ref{tab8}. 
On this subset, we set the tolerance level $\alpha=0.1$ to rigorously calibrate the raw attribute prediction score thresholds from RemoteCLIP-FG.
However, statistical calibration requires sufficient sample density to be reliable. 
To address corner cases where specific attribute categories are either absent or sparsely represented (fewer than 10 instances) in the calibration set, we bypass the quantile calculation. 
Instead, we impose a conservative fixed threshold (e.g., $\tau=0.2$) for these long-tail attributes. 
This fallback strategy is designed to prioritize precision, effectively suppressing potential false positives that could arise from unstable statistical estimates on scarce data. 

\begin{table*}[h]
    \centering
    \small
    \caption{Details of the calibration dataset with attribute statistics for each category.}
    \label{tab8}
    \begin{tabular*}{\textwidth}{@{\extracolsep{\fill}} l l}
        \toprule
        Category & Details(The numbers represent the quantity) \\
        \midrule
        \multirow{4}{*}{Plane}
        & \multirow{4}{*}{%
            {\tiny
            \begin{tabular}[l]{@{}l@{}}
                \textbf{Number of Images:}: 
                4882. \\
                \textbf{ProtoTag}: 
                Plane (4882). 
                \textbf{Propulsion type}: 
                Jet (3820);
                Propeller (2429). 
                \textbf{Number of engines}: 
                One-engine (689); 
                Eight-engine (275);
                Four-engine (2616);
                Twin-engine (2669).  \\
                \textbf{Engine position}: 
                One the nose (523); 
                At wing roots and lower fuselage (1224);
                Beneath the wings (3866); 
                Embedded within wing (15); 
                Above the wings (2); 
                Rear fuselage (621). \\ 
                \textbf{Purpose}: 
                Aerial Support Aircraft (688); \\
                Airborne Early Warning Aircraft (478); 
                Airline Aircraft (580); 
                Anti-Submarine Warfare Aircraft (518);
                Bomber (698); 
                Trainer (556); 
                Chartered aircraft (501);
                Attack aircraft (351);  \\
                Fighter (620);
                Transport Aircraft (699).
                \textbf{Usage}: 
                Civilian Aircraft (1617);
                Commercial Aircraft (581);
                Military Aircraft (4027). 
                \textbf{Wing configuration}: 
                Straight wing (2208); \\
                Swept diamond-like wing (85); 
                Flying wing (15); 
                Swept delta wing (185); 
                Swept wing (3565);
                Swept, variable-sweep wing (373).
            \end{tabular}}
        } \\
        &  \\
        &  \\
        &  \\
        &  \\
        \midrule
        \multirow{2}{*}{Ship}
        & \multirow{3}{*}{%
            {\tiny
            \begin{tabular}[l]{@{}l@{}}
                \textbf{Number of Images:}: 
                9028. \\
                \textbf{ProtoTag}: 
                Ship (9028). 
                \textbf{Purpose}: 
                Aircraft Carrier (24);
                Amphibious Ship (30);
                Auxiliary Ship (6);
                Cargo Ship (2789);
                Commander (20);
                Cruiser (32); \\
                Destroyer (698);
                Fishing Vessel (1283);
                Frigate (142);
                Medical Ship (4);
                Military Transport Ship (9);
                Motorboat (3626); 
                Passenger Ship (519);
                Patrol (60); \\
                Submarine (202);
                Landing (23); 
                Medical Ship (13); 
                Tugboat (323). 
                \textbf{Subcat}: 
                Barge (668);
                Container Ship (301);
                Cruise Ship (22);
                Dry Cargo Ship (1014);
                Liquid Cargo Ship (789); \\
                RoRo (17); 
                Yacht (497). 
                \textbf{Usage}: 
                Civilian Ship (5428);
                Commercial Ship (2789);
                Engineering Ship (323);
                Military Ship (488).
            \end{tabular}}
        } \\
        &  \\
        &  \\
        &  \\
        \midrule
        \multirow{1}{*}{Vehicle}
        & \multirow{1}{*}{%
            {\tiny
            \begin{tabular}[l]{@{}l@{}}
                \textbf{Number of Images:}: 
                5360. \\
                \textbf{ProtoTag}: 
                Vehicle (5360). 
                \textbf{Purpose}: 
                Bus (560);
                Cargo Truck (651);
                Dump Truck (562);
                Excavator (696);
                Pick-up (658);
                Small Passenger Car (705); \\
                Tractor (349); 
                Truck Tractor (525);
                Van (654). 
                \textbf{Usage}: 
                Engineering Vehicle (1045);
                Large Civilian Vehicle (560);
                Small Civilian Vehicle (2017);
                Truck (1738).
            \end{tabular}}
        } \\
        &  \\
        \bottomrule
    \end{tabular*}
\end{table*}

\subsection{Scale Stage.} 
In this stage, we aim to scale up the fine-grained supervision.
First, we curate a self-built unlabeled dataset, referred to as RS-C.
To equip RS-C with object localization information, we utilize a detector trained on the DOTA-v2.0 training set to generate horizontal bounding box annotations.
Prioritizing precision over recall to minimize noise, we employ the DOTA-v2.0 validation set as the calibration set.
By setting the conformal tolerance level to $\alpha=0.1$, we derive rigorous class-wise score thresholds to filter these bounding box predictions, ensuring high-quality localization.
Subsequently, we deploy the RemoteCLIP-FG teacher model, alongside its calibrated attribute thresholds (derived in the Calibrate Stage), to perform attribute pseudo-labeling and filtering on both the aggregated public datasets (RS-O) and the newly curated RS-C.
This process culminates in the creation of two massive attribute-annotated datasets: RS-Attribute-O and RS-Attribute-C, which serve as the data foundation for our SLIP-RS. 

\section{Fine-grained and Attribute Recognition Benchmark} 

We construct a specialized benchmark derived from the authoritative datasets, including Fair1M~\cite{sun2022fair1m}, MAR20~\cite{wenqi2024mar20}, and HRSC2016. 
This benchmark encompasses three categories (plane, ship, and vehicle) and is meticulously annotated with hierarchical attributes, ranging from high-level functional descriptions (e.g., \textit{Purpose}, \textit{Usage}) to explicit physical components (e.g., \textit{Wing Configuration}, \textit{Engine Position}). 
The detail is shown in Table~\ref{tab9}, and the processing details are listed below: 
\begin{itemize}
    \item MAR20: We utilize both the validation and test sets. Attribute annotations are derived by mapping the specific aircraft model labels to their corresponding structural attributes.
    \item HRSC2016: We employ the entire dataset for construction to maximize the diversity of ship attributes, utilizing the existing fine-grained class labels for attribute derivation. 
    \item Fair1M: We select the validation set and preprocess the high-resolution imagery by cropping it into non-overlapping $1024 \times 1024$ patches. Regarding attribute generation, we adopt a category-specific strategy: for Airplane instances, we derive attributes via explicit mapping from fine-grained aircraft models; conversely, for vehicle and ship categories, we directly inherit the original fine-grained annotations to construct the attribute set. 
\end{itemize}

\begin{table*}[t]
    \centering
    \small
    \caption{Details of the fine-grained attribute benchmark, which summarizes the data sources and attribute statistics for each category(`P' for Purpose, `U' for Usage, `EP' for Engine Position, `WC' for Wing Configuration, `NE' for Num of Engines, `PT' for Propulsion Type, `S' for Subcat).} 
    \label{tab9}
    \begin{tabular*}{\textwidth}{@{\extracolsep{\fill}} l l l}
        \toprule
        Category & Source & Details(The numbers represent the quantity) \\
        \midrule
        \multirow{3}{*}{Plane}
        & \multirow{2}{*}{Fair1m-val}
        & \multirow{3}{*}{%
            {\tiny
            \begin{tabular}[l]{@{}l@{}}
                \textbf{Number of Images:}: 
                3394. 
                \textbf{Number of Instances}: 
                15294; \\
                \textbf{ProtoTag}: 
                Plane (15294). 
                \textbf{Propulsion type}: 
                Jet (14013);
                Propeller (1281). 
                \textbf{Number of engines}: 
                Eight-engine (296);
                Four-engine (4075);
                Twin-engine (10923).  \\
                \textbf{Engine position}: 
                At wing roots and lower fuselage (864);
                Beneath the wings (11594);
                Rear fuselage (2836). 
                \textbf{Purpose}: 
                Aerial Support Aircraft (267); \\
                Airborne Early Warning Aircraft (221); 
                Airline Aircraft (8756); 
                Anti-Submarine Warfare Aircraft (350);
                Bomber (1157); 
                Chartered aircraft (133);
                Fighter (2706); \\
                Transport Aircraft (1704).
                \textbf{Usage}: 
                Civilian Aircraft (133);
                Commercial Aircraft (8756);
                Military Aircraft (6405). 
                \textbf{Wing configuration}: 
                Straight wing (922); \\
                Swept diamond-like wing (181);
                Swept wing (12675);
                Swept, high aspect-ratio wing (655);
                Swept, variable-sweep wing (861).
            \end{tabular}}
        } \\
        &       &  \\
        & MAR20- &  \\
        & val\&test      &  \\
        \midrule
        \multirow{2}{*}{Ship}
        & \multirow{2}{*}{Fair1m-val}
        & \multirow{3}{*}{%
            {\tiny
            \begin{tabular}[l]{@{}l@{}}
                \textbf{Number of Images:}: 
                2951. 
                \textbf{Number of Instances}: 
                10558; \\
                \textbf{ProtoTag}: 
                Ship (10558). 
                \textbf{Purpose}: 
                Aircraft Carrier (111);
                Amphibious Ship (141);
                Auxiliary Ship (16);
                Cargo Ship (4570);
                Commander (54);
                Cruiser (142); \\
                Destroyer (240);
                Fishing Vessel (1087);
                Frigate (156);
                Medical Ship (4);
                Military Transport Ship (90);
                Motorboat (3181); 
                Passenger Ship (463);
                Patrol (2); \\
                Submarine (74);
                Tugboat (227). 
                \textbf{Subcat}: 
                Barge (11);
                Container Ship (51);
                Cruise Ship (47);
                Dry Cargo Ship (3843);
                Liquid Cargo Ship (597);
                RoRo (68); \\
                Yacht (416). 
                \textbf{Usage}: 
                Civilian Ship (4731);
                Commercial Ship (4570);
                Engineering Ship (227);
                Military Ship (1030).
            \end{tabular}}
        } \\
        &          &  \\
        & HRSC2016 &  \\
        &          &  \\
        \midrule
        \multirow{1}{*}{Vehicle}
        & \multirow{1}{*}{Fair1m-val}
        & \multirow{1}{*}{%
            {\tiny
            \begin{tabular}[l]{@{}l@{}}
                \textbf{Number of Images:}: 
                2253. 
                \textbf{Number of Instances}: 
                132162; \\
                \textbf{ProtoTag}: 
                Vehicle (132162). 
                \textbf{Purpose}: 
                Bus (611);
                Cargo Truck (7035);
                Dump Truck (1605);
                Excavator (370);
                Pick-up (5456);
                Small Passenger Car (56149); \\
                Tractor (36); 
                Truck Tractor (358);
                Van (60542). 
                \textbf{Usage}: 
                Engineering Vehicle (406);
                Large Civilian Vehicle (611);
                Small Civilian Vehicle (122147);
                Truck (8998).
            \end{tabular}}
        } \\
        &          &  \\
        \bottomrule
    \end{tabular*}
\end{table*}

\section*{Limitations and Future Work}

Our structured attribute construction is highly sophisticated for \textit{Planes}, covering intricate details like engine layout and wing shape. 
In contrast, the attribute definitions for \textit{Ships} and \textit{Vehicles} remain relatively coarse. 
Future work will aim to bridge this gap by extending the paradigm to capture more complex structural semantics, such as \textit{island position} and \textit{mast count} for ships, as well as integrating universal visual primitives like \textit{color} and \textit{scale}. 
We believe this expansion will drive the model to learn highly discriminative fine-grained representations, thereby fortifying its universality across broader remote sensing scenarios. 

Furthermore, we have observed preliminary evidence that learning shared attributes endows the model with intrinsic compositional reasoning capabilities. 
This opens up a promising avenue for detecting unseen categories simply by permuting and recombining known attribute primitives. 
We intend to rigorously explore this direction, leveraging the combinatorial logic of attributes to achieve true compositional zero-shot detection in open-world scenarios. 

\end{document}